\documentclass[10pt,twocolumn,letterpaper]{article}

\usepackage[pagenumbers]{cvpr} 

\definecolor{cvprblue}{rgb}{0.21,0.49,0.74}
\pdfoutput=1
\usepackage[pagebackref,
breaklinks,
colorlinks, 
citecolor=cvprblue,
linkcolor=red,
anchorcolor=red,
]{hyperref}

\def\paperID{8832} 
\def\confName{CVPR}
\def\confYear{2025}

\title{SpecDM: Hyperspectral Dataset Synthesis with Pixel-level Semantic Annotations}

\author{Wendi Liu\thanks{Equal contribution}\qquad
Pei Yang\footnotemark[1]\qquad
Wenhui Hong\qquad
Xiaoguang Mei\thanks{Corresponding author}\footnotemark[2]\qquad
Jiayi Ma\\
Electronic Information School, Wuhan University\\
{\tt\small \{lwd2018\_360, 2019yp\}@whu.edu.cn, Wenhui\_Hong@foxmail.com, \{meixiaoguang, jyma2010\}@gmail.com}}

\begin{document}
\maketitle
\begin{figure*}[t]
    \centering
    \includegraphics[width=\linewidth]{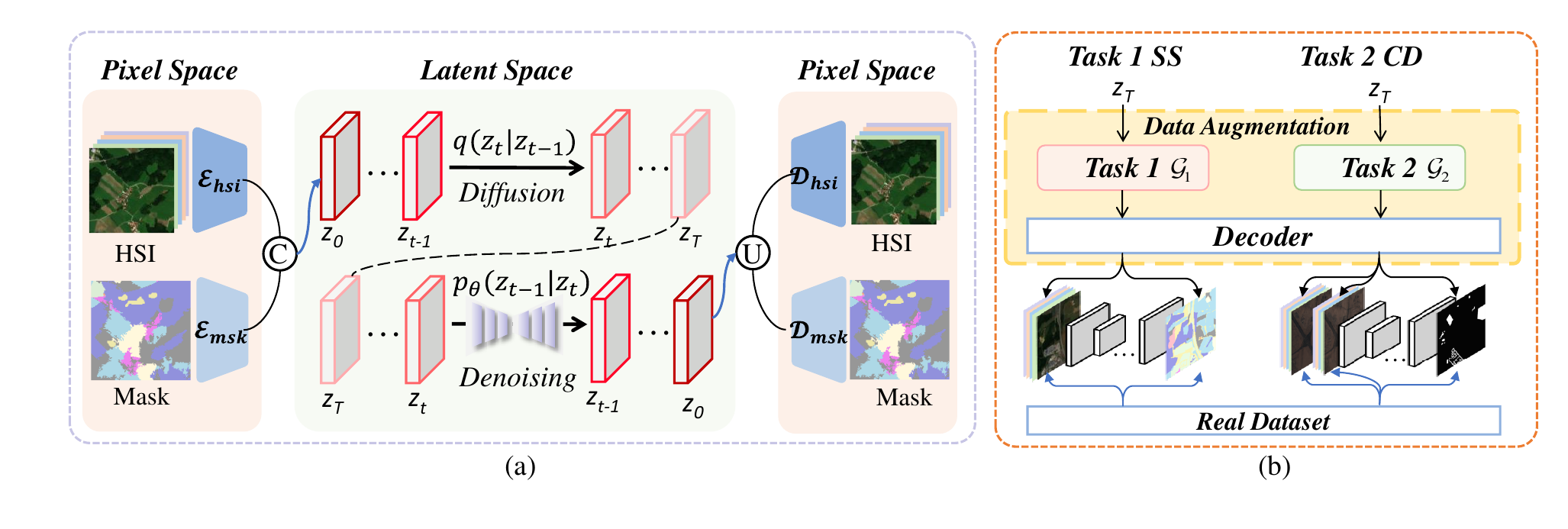}
    \vspace{-10mm}
    \caption{\textbf{Overview of our approach.} (a) In the \textbf{training} stage, we design a two-stream VAE to compress HSIs and corresponding masks from pixel space to latent space, and then train a denoising U-Net on the joint representations. The latent representation is split to feed forward to corresponding decoders to complete the reconstruction. (b) In the \textbf{inference} stage, after training the generator $\mathcal G$, we start from the noised sample $z_T$ and obtain the synthetic image-mask pairs through decoders, to augment the original real dataset when training the downstream task models.}
    \label{fig:approach}
\end{figure*}
\begin{abstract}
In hyperspectral remote sensing field, some downstream dense prediction tasks, such as semantic segmentation (SS) and change detection (CD), rely on supervised learning to improve model performance and require a large amount of manually annotated data for training. However, due to the needs of specific equipment and special application scenarios, the acquisition and annotation of hyperspectral images (HSIs) are often costly and time-consuming. To this end, our work explores the potential of generative diffusion model in synthesizing HSIs with pixel-level annotations. The main idea is to utilize a two-stream VAE to learn the latent representations of images and corresponding masks respectively, learn their joint distribution during the diffusion model training, and finally obtain the image and mask through their respective decoders. To the best of our knowledge, it is the first work to generate high-dimensional HSIs with annotations. Our proposed approach can be applied in various kinds of dataset generation. We select two of the most widely used dense prediction tasks: semantic segmentation and change detection, and generate datasets suitable for these tasks. Experiments demonstrate that our synthetic datasets have a positive impact on the improvement of these downstream tasks.
\end{abstract}    
\section{Introduction}
\label{sec:intro}
Hyperspectral image, with its 3D data structure, provides more detailed spectral information compared to RGB image, which makes it take advantages in various applications such as face recognition \cite{face1, face2, face3}, vegetation detection \cite{vegetation1, vegetation2, vegetation3} and geological observation \cite{earth1, earth2, earth3}. However, owing to the performance of the equipments, the requirements of scenes and objects, and the limitations of the environment, it is costly to obtain the HSI data \cite{scarse1, scarse2}. For some visual tasks with dense prediction, the cost of label annotation cannot be ignored either, especially for remote sensing scenes with large fields. In addition to the cost of annotation, the sensitivity of some hyperspectral data also makes it difficult for ordinary researchers to access the data. Due to the above reasons, both the construction of large-scale hyperspectral dataset platforms, and the research of data-dependent AI models in the field of HSI are currently severely hindered \cite{challenge1, challenge2}.
To address the scarcity of HSI data, some researchers usually use techniques such as affine transformation to enhance data \cite{affine}, or use physical modeling based synthetic data \cite{physics}. Some research also explore to reconstruct the spectral information from RGB images \cite{he2023spectral}. However, such techniques either fail to substantially increase the diversity of data, or produce high-quality data limited by the physical model.

Recently, generative AI models, such as Variational Autoencoder (VAE) \cite{vae1, vae2}, Generative Adversarial Network (GAN) \cite{gan1, gan2} and Diffusion Model (DM) \cite{dm1, dm2, ldm}, have achieved great success in the field of natural image synthesis. In most visual tasks, especially supervised learning, high-quality data annotation is as significant as the image data itself. While working on generating images with rich visual effects, some works are also devoted to exploring the generation of annotated datasets \cite{datasetgan, bigdatasetgan, diffumask, datasetdm}. For example, DatasetDM \cite{datasetdm} designed a unified perception decoder which can generate different perception annotations to meet the demands of various downstream tasks. In optical remote sensing field, SatSynth \cite{satsynth} used DDPM \cite{dm2} to generate images and segmentation masks simultaneously. For HSI synthesis, it is difficult to automatically generate such annotations through algorithms since most existing dense prediction methods like SAM \cite{sam} are designed for RGB images and cannot be directly applied to high-dimensional HSIs. Hence existing research is still at the stage of pure image generation \cite{yu2024unmixing, yu2024unmixdiff, hsigene} and cannot meet the demands of downstream tasks which need pixel-level annotations.

In this work, we focus on filling the gap in the field of hyperspectral data generation, exploring the potential of diffusion model to augment existing hyperspectral datasets in a generative manner. In addition to image data, our work can also simultaneously generate semantic labels suitable for downstream dense prediction tasks, specifically, for semantic segmentation and change detection, which are two significant tasks in hyperspectral remote sensing field. To the best of our knowledge, it is the first work to generate high-dimensional HSIs with pixel-level annotations. Instead of additionally designing a segmentation or change detection algorithm to generate annotations of HSIs, our work directly learn the joint distribution of image-label pairs by designing a two-stream training paradigm for the first-stage training, based on the classic Latent Diffusion Model (LDM) \cite{ldm}. Specifically, we implement a two-stream variational autoencoder, corresponding to the image data stream and the label data stream respectively. Due to the different distribution between the HSI pixel value and mask value \cite{satsynth}, the two VAE branches use different network parameters. In the second-stage diffusion and denoising process, we concatenate the latent features of image and semantic mask in the channel dimension and to learn their joint distribution. When generating, we sample from the joint distribution to get latent codes, then decouple and decode them to obtain high-quality images and semantic labels separately.

To summarize, the contributions of our work are as below:

\begin{itemize}
    \item We propose SpecDM: a new dataset synthesis method for hyperspectral images utilizing the generative diffusion model, which can generate high quality training data instances with pixel-level semantic labels.

    \item In order to solve the distribution difference between image and label value domains, we design the two-stream VAE to separately learn the latent representation of image and label. In addition to semantic segmantation, we expand this training paradigm to change detection. 

    \item Experiments demonstrate that the existing models trained on augmented data generated by our method exhibit significant improvements on semantic segmentation and change detection, which are two main downstream tasks in hyperspectral remote sensing fields.
\end{itemize}
\section{Related Work}
\label{sec:related work}

\subsection{Generative AI-based Data Synthesis}

Recently, many mainstream data synthesis methods have relied on generative AI models, including VAE-based \cite{shi2019variational,  Child2020VeryDV,vahdat2020nvae}, GAN-based \cite{Example-Guided, Controllable_Person, esser2021taming}, and DM-based \cite{kim2023dcface,zhang2023adding,li2024generative,ma2024latte} methods. With the emergence of large generative models such as DALLE-3, Stable Diffusion 3, and Sora, synthetic images and videos have achieved astonishing visual effects regarding diversity and authenticity.

In addition to merely use generative models to synthesize visually appealing images, previous works\cite{gaidon2016virtual,kar2019meta,devaranjan2020meta} have leveraged 3D graphics engines to generate labeled datasets. However, The scene diversity and authenticity of these synthetic datasets are still very limited. To make the scene more realistic, some studies \cite{datasetgan,bigdatasetgan} focus on GAN-based models to produce images via image translation to avoid the domain gap brought by graphics rendering. Inspired by the success of the diffusion model in image generation, recent work has begun to explore its potential in dataset synthesis with pixel-wise labels. DiffuMask \cite{diffumask} automatically obtains synthesized images and semantic masks through text-driven diffusion models. To accommodate various downstream tasks, DatasetDM \cite{datasetdm} employs a pre-trained diffusion model with a multi-task decoder to synthesize different perception annotations. Different from existing generative models designed for RGB data, our work focuses on the generation of higher-dimensional HSI data in the field of remote sensing.


\subsection{Hyperspectral Data Synthesis}
Due to the high-dimensional characteristics of HSI data, generating large-scale datasets has always been an extremely challenging task. Previous works can be roughly divided into three categories: physical simulation based on imaging systems \cite{physics,verrelst2015optical}, augmentation based on affine transformation \cite{affine,zhang2023features,wang2023multistage}, and spectral super-resolution reconstruction \cite{he2023spectral,cai2022mask,cai2022mst++}. These methods provide a feasible solution to the persistent data shortage, while they can not produce truly new samples. 

More recently, some explorers have introduced Diffusion models into HSI data synthesis.
Considering the spectral properties, UnmixDiff \cite{yu2024unmixdiff} has performed the diffusion process in the abundance domain of HSI. Unmixing Before Fusion \cite{yu2024unmixing} has gone one step further and designed a pipeline for synthesizing HSI that couples the multi-source unmixing model and diffusion model, utilizing rich RGB images to guide the model to learn the spatial distribution characteristics of real scenes and improve the diversity of generation. To obtain more precise and reliable HSI data, HSIGene \cite{hsigene} has employed LDM with multiple control conditions. Meanwhile, to enhance the spatial diversity, HSIGene has appended a super-resolution model to achieve data augmentation after the generation. However, the synthetic data obtained by the mentioned approaches above is only suitable for tasks that do not require annotation costs (such as denoising and super-resolution) and some downstream tasks with low manual annotation costs, such as scene classification, which only requires image-level annotations. Different from existing approaches, our work firstly generates joint pairs of HSI data with pixel-wise labels, which can be applicable in dense perception task predictions, such as semantic segmentation and change detection.

\subsection{Semantic Segmentation and Change Detection}
Semantic segmentation and change detection are typical tasks in the field of HSI remote sensing understanding. The former aims to assign a semantic category to each pixel of an image, while the latter aims to detect changes in objects by using images in different time phases. Compared with natural image datasets, HSI satellite images face unique dilemmas \cite{he2017recent,li2019deep}: relatively small training set compared to the high-dimensional spectra, which adversely affects the performance of segmentation and detection models. 

To address such challenges, many deep learning-based methods \cite{chen2016deep, sun2019hyperspectral} are dedicated to exploring dimensionality reduction or band selection techniques to reduce the impact of redundant information. Although significant progress has been made, the development of these two tasks is severely restricted by the availability of HSI data \cite{li2019deep,liu2019review}. 
To alleviate the pressure of annotation, some works \cite{10354413, Gao_2021_CVPR,manas2021seasonal} have explored unsupervised learning, but the performance has significantly declined. Therefore, we propose to directly generate joint image-label pairs through the generative model and verify the effectiveness of the synthetic datasets in improving the accuracy of semantic segmentation and change detection.
\section{Method}
\label{sec:method}

Our proposed SpecDM comprises two stages, which is illustrated in Fig. \ref{fig:approach}. The \textbf{Training} stage involves compressing the data through the two-stream VAE to obtain the latent representations of image and semantic label separately, and learning the mapping from Gaussion distribution to the joint distribution of image-mask pairs by training a denoising U-Net \cite{unet}. In the \textbf{Inference} stage, we sample the joint latent representations from Gaussian distribution and denoise it through the denoising U-Net. The clean latent representation is then decomposed to the image and label parts, and decoded by the corresponding decoders to obtain HSIs and annotations.

\subsection{Two-stream Encoding for Data Compression}

Due to the high-dimensional spectral information of HSI, training DMs in original image space is computationally expensive. Previous works using unmixing to map the HSI to the low-dimensional abundance space to ensure the fidelity of spectral response of synthetic HSI \cite{yu2024unmixdiff, yu2024unmixing}. While generating high-quality HSI, such a compression approach faces two challenges: (i) The dimension of unmixing is corresponded to the number of endmembers. When the dataset covers a larger variety of materials, the dimension of abundance is still high after unmixing, which is not suitable for segmentation datasets with more types of landforms. (ii) As a dimension reduction method, unmixing cannot handle the low-dimensional annotation images, such as binary masks. In this case, forced unmixing will lose its original physical meaning.

In this work, we propose to use two-stream encoding for data compression. Specifically, two branches of VAE in original LDM are used to encode the input data pairs, while one branch is used to encode the image data, and the other branch encodes the annotation data. Given an HSI $x\in\mathbb{R}^{H\times W\times C}$ with the semantic mask $y\in\mathbb{Z}^{H\times W}$, the image branch encoder $\mathcal{E}_{hsi}$ and mask branch encoder $\mathcal{E}_{msk}$ encode $(x, y)$ pairs into the joint latent representations $(z_x, z_y) = (\mathcal{E}_{hsi}(x), \mathcal{E}_{msk}(y))$, where $z_x, z_y \in\mathbb{R}^{h\times w\times c}$. Downsampling factor is defined as $f = H/h = W/w$. The decoders $\mathcal{D}_{hsi}$ and $\mathcal{D}_{msk}$ reconstructs the image and mask from $(z_x, z_y)$ pairs. To reconstruct HSI, we add the spectral angle distance (SAD) measurement as a part of loss function in additional to original loss to ensure the spectral fidelity. Then the loss function of image branch $\mathcal L_{hsi}$ is defined as:
\begin{equation}
\label{eq:image loss}
    \mathcal L_{hsi} (x, \hat{x}) = \mathcal L_1(x, \hat{x}) + \lambda \arccos(\frac{x\hat{x}^T}{\Vert x\Vert_2\Vert \hat{x}\Vert_2}),
\end{equation}
where $\mathcal L_1$ represents the $L_1$ loss and $\lambda$ is used to balance the two items. To reconstruct semantic mask, we use cross entropy loss, then the total loss of the two-stream VAE is defined as:
\begin{equation}
    \mathcal{L} = \mathcal L_{hsi}(x, \hat{x}) +\mathcal L_{CE} (y, \hat{y}),
\end{equation}

It should be noted that the two branches have totally different parameters for the great difference between continuous image pixel values in $\mathbb{R}$ and discrete mask values in $\mathbb{Z}$. In this manner, we can perform image and annotation data compression simultaneously without being constrained by the form of unmixing. In order to take into account both the computational efficiency in the subsequent diffusion process and reconstruction quality, we choose a downsampling factor $f = 4$ \cite{ldm}.

\subsection{Diffusion Model of Joint Representations}

After getting the joint latent representations $z = (z_x, z_y)$ of image and semantic mask inputs in the first-stage, we approximates the posterior distribution $q(z_{1:T}|z_0)$ through the diffusion forward process, and then training the denoising U-Net to denoise from $p_\theta(z_{t-1}|z_t)(t=\{1, ..., T\})$ to obtain the clean reconstruction step by step. Here we provide a brief introduction of this process.

Given a joint latent representation $z = (z_x, z_y) \in\mathbb R^{h\times w\times 2c}$, the diffusion process gradually adds Gaussian noise following a pre-defined noise schedule $\beta_1, . . . , \beta_T$:
\begin{equation}
    q(z_t|z_{t-1}):=\mathcal N(z_t;\sqrt{1-\beta_t}z_{t-1}, \beta_t\mathrm{\mathbf{I}}),
\end{equation}
where $t$ represents the $t$-th time step. After sufficiently large $T$ steps, we obtain a Gaussion random noise sample $z_T\sim \mathcal N(0, \mathrm{\mathbf{I}})$. The reversed denoising process is performed through the U-Net by optimizing the following objective function:

\begin{equation}
    \mathcal L=\mathbb E_{z, \epsilon\sim\mathcal N(0, \mathrm{\mathbf{I}}), t}\left[\Vert {\epsilon - \epsilon_\theta(z_t, t)}\Vert_2^2\right],
\end{equation}
Thus we have completed the reconstruction from Gaussian distribution to the input training data distribution.

\subsection{Data Synthesis with Semantic Annotation}

In this work, we preset two types of dataset generation tasks, one for semantic segmentation and the other for change detection, which are two typical pixel-level dense prediction tasks.

\noindent
\textbf{Synthesis for Semantic Segmentation.}  To synthesize image-mask pairs for semantic segmentation, we take the following steps:

\begin{itemize}
    \item Train the two-stream VAE on data pairs $(x, y)$ to get the joint latent representations $z = (z_x, z_y)$.
    \item Train the diffusion model $\mathcal G$ in the latent space.
    \item Sample from $\mathcal G$ to get synthetic latent representations $z_{syn}$.
    \item Decode $z_{syn}$ using the trained decoders $\mathcal D_{hsi}$ and $\mathcal D_{msk}$ to get synthetic pairs $(x_{syn}, y_{syn})$.
\end{itemize}

\noindent
\textbf{Synthesis for Change Detection.}  Such paradigm can be expanded to change detection dataset synthesis. For change detection, a data instance consists of two images at different temporal phases and a mask to represent the change. While expanding to change detection, the mask branch keeps the same, and the image branch accepts the two images as inputs. Since the image branch is utilized to compress image data only, there is no need to add additional branches with different parameters even if the the interface for input images is increased. In this case, the inputs is encoded as $z = (z_{x_1}, z_{x_2}, z_y)\in\mathbb R^{h\times w\times 3c}$, where $(z_{x_1}, z_{x_2}) = \mathcal E_{hsi}(x_1, x_2)$ and $z_y = \mathcal E_{msk}(y)$.

\subsection{Implementation Details}
\noindent
\textbf{Latent Diffusion.} We follow the LDM \cite{ldm} to set our experiments configurations. For two-stream VAE training, we take KL-regularized VAE as the backbone of both image and mask branches. Image branch accepts multi-channel HSIs as inputs, and the mask branch accepts one-hot encodings as inputs.
The SAD tradeoff $\lambda$ in Eq. (\ref{eq:image loss})  is set to 0.1, and the initial learning rate is set to $4.5\times 10^{-6}$. For diffusion model training, we apply $T=1000$ denoising steps with a linear $\beta$ schedule from 0.0015 to 0.0155. The learning rate is set to $5.0\times 10^{-6}$.

\noindent
\textbf{Downstream Task.} For \textbf{semantic segmentation}, we choose SegFormer \cite{segformer} and PFSegNet \cite{pfsegnet} algorithms to evaluate the performance trained on the original dataset and augmented dataset, respectively. Since the SegFormer was designed for RGB semantic segmentation, we add a mapping layer before the backbone to map the input HSI to 3 channels and load the pre-trained backbone model. For \textbf{change detection}, we use SiamCRNN \cite{siamcrnn} and ChangeFormer \cite{changeformer} algorithms to evaluate the performance.

All of above experiments were carried out using 4 NVIDIA 3090 GPUs.
\label{sec:syn performance}
\begin{figure*}[t]
    \centering
    \includegraphics[width=1.0\linewidth]{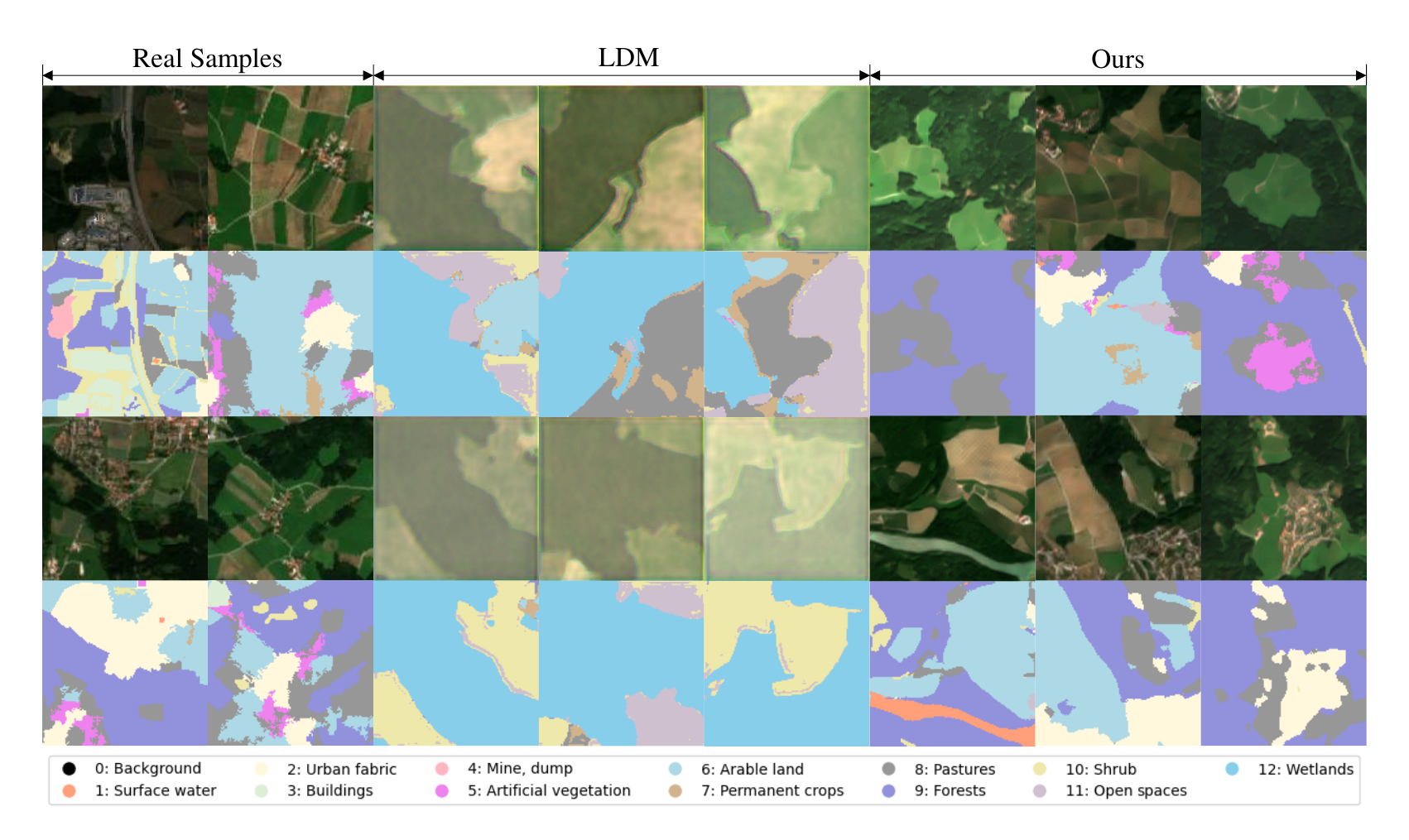}
    \vspace{-5mm}
    \caption{\textbf{Generated samples-SegMunich.} We visualize several pairs of HSIs (shown in false-color) and corresponding segmentation maps generated by the baseline method LDM and our method respectively, comparing to the real samples.}
    \label{fig:Visual Quality-SS}
\end{figure*}
\section{Experiments}
\label{sec:experiments}
\subsection{Datasets}
\noindent
\textbf{SegMunich.} The SegMunich dataset is selected to perform the semantic segmentation data synthesis and the downstream task. This dataset, captured in Munich's urban from Sentinel-2 spectral satellite,  was first created and utilized in the published work SpectralGPT \cite{spectralgpt}. It consists of 13 bands with a spatial resolution of 10 meters, including the segmentation mask that meticulously delineates 13 Land Use and Land Cover (LULC) classes. The original work \cite{spectralgpt} chooses to combine the 10-meter spectral bands (B1, B2, B3, and B4) with resampled 20-meter spectral bands (B5, B6, B7, B8A, B11, B12) to get the 10-bands patches, to create a comprehensive feature representation for semantic segmentation. Our work keeps the same band configuration. The original dataset consists 39402 pairs for training and 9846 pairs for validation. We removed the patches which contain a lot of blank background (e.g., the entire image is occupied by the blank background). The cleaned dataset has 21680 pairs for training and 5410 pairs for validation with a patch size $128\times 128$.

\noindent
\textbf{OSCD.} The Onera Satellite Change Detection (OSCD) dataset \cite{oscd} is utilized to perform the change detection data synthesis and the downstream task. This dataset comprises 24 cities of Sentinel-2 images, captured between 2015 and 2018. The original images have 13 bands. Since the OSCD dataset is captured by the same satellite as the SegMunich dataset, we select the same bands combination as SegMunich for convenience. The images and masks are cropped to 237 pairs for training and 86 pairs for validation, with a 60\% overlap rate and a patch size $256 \times 256$.

\begin{figure*}[t]
    \centering
    \includegraphics[width=1.0\linewidth]{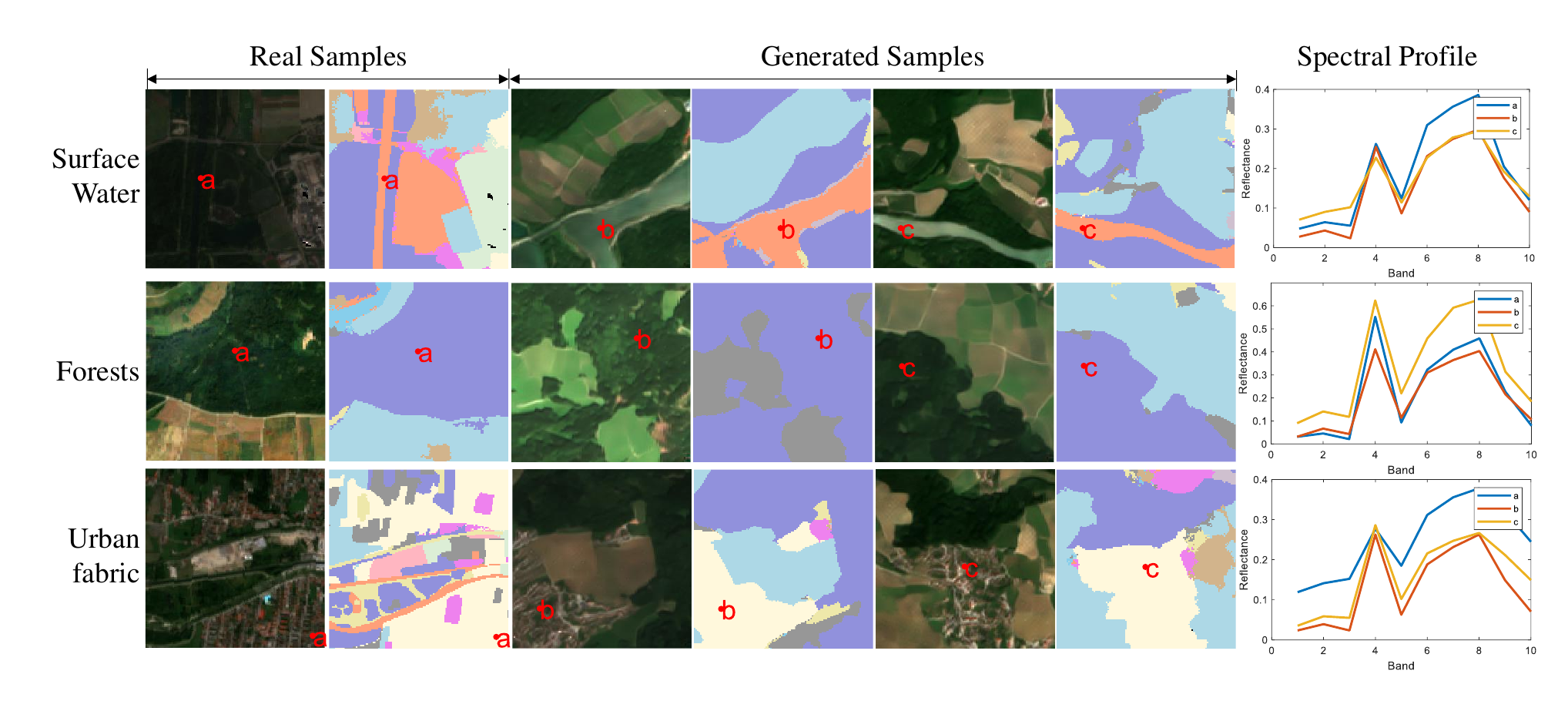}
    \vspace{-10mm}
    \caption{\textbf{Spectral profile comparison.} We visualize the spectral response of our generated samples, comparing to real samples. We sample the pixels of several typical landforms according to the annotations. The intensity of spectral responses of the same landform keep consistent in different HSIs and are close to the real samples.}
    \label{fig:Spectral analysis}
\end{figure*}

\subsection{Synthesis Performance}
\noindent
\textbf{Visual Quality.} We utilized typical LDM \cite{ldm} as the baseline to evaluate the sample quality of our synthesis method. In the first stage training, we simply concatenate the image and mask in the channel dimension to get the input. Hence LDM can be regarded as encoding the image and mask using only single-stream VAE. We use Frechet Inception Distance (FID) \cite{fid} to measure the similarity of distributions of real dataset and synthetic dataset. The comparison results are displayed in Table \ref{tab:visual quality-ss}, confirming the superior visual quality of our generated samples.

\begin{table}[t]
  \centering
    \begin{tabular}{lccc}
    \toprule
    Method & FID (Image)$\downarrow$ & FID (Mask)$\downarrow$ & mSAD$\downarrow$ \\
    \midrule
    LDM-SS & 70.44 & 50.01 & 0.13\\
    Ours-SS & 5.19 & 10.79 & 0.03\\
    \bottomrule
  \end{tabular}
  \caption{\textbf{Quantitative evaluation of synthetic dataset-SegMunich.} The first two columns display the FID scores of image and label respectively. The last column displays the mSAD scores to evaluate the spectral fidelity of generated samples. In both tables, ($\downarrow$) indicates lower metric values are better, whereas ($\uparrow$) denotes higher values are better.}
  \label{tab:visual quality-ss}
\end{table}

\begin{figure}[t]
    \includegraphics[width=1.0\linewidth]{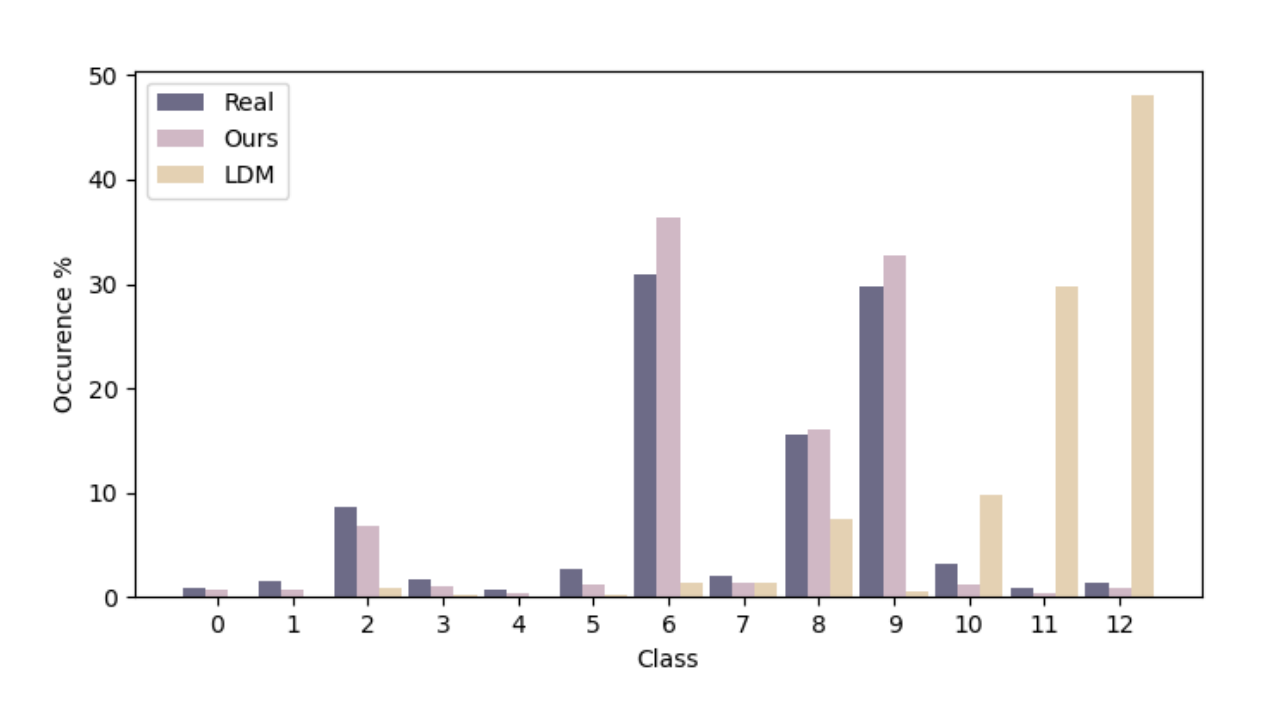}
    \vspace{-7mm}
    \caption{\textbf{Distribution of landform classes,} illustrating that the set of our generated samples closely matches the real distribution. The proportion of several main classes are very close (Arable land, Pastures and Forests).}
    \label{fig:Distribution of landform classes}
\end{figure}

We further provide qualitative samples of generated training pairs in Fig. \ref{fig:Visual Quality-SS}, compared to LDM method, and the distribution of landform classes in Fig. \ref{fig:Distribution of landform classes}. We can observe that the samples generated by our method have the similar spatial distribution with the real dataset. The edges of landforms in image also have great consistency with the mask. The proportion of main types of landforms, such as Arable land, Pastures and Forests, is close to the real dataset. On the contrary, the samples generated by LDM have a large deviation from the original real dataset. The proportion of different types of landforms also shows a large difference from the real data.

\noindent
\textbf{Spectral Fidelity.} Since this work focuses on the spectral data synthesis, the quality of generated spectra is essential in evaluation. We calculate the average spectral response of each class of landform, and compare the mean SAD with the real data. The results are displayed in the last column in Table \ref{tab:visual quality-ss}, which illustrate that the spectra of each class generated by our method is close to the real dataset. We also display the spectral profiles of several typical landforms, sampled from real and our generated samples, showed in Fig. \ref{fig:Spectral analysis}. Our generated samples exhibit strong spectral consistency with the real data for the same landform.

\begin{figure*}[th]
    \centering
    \includegraphics[width=1.0\linewidth]{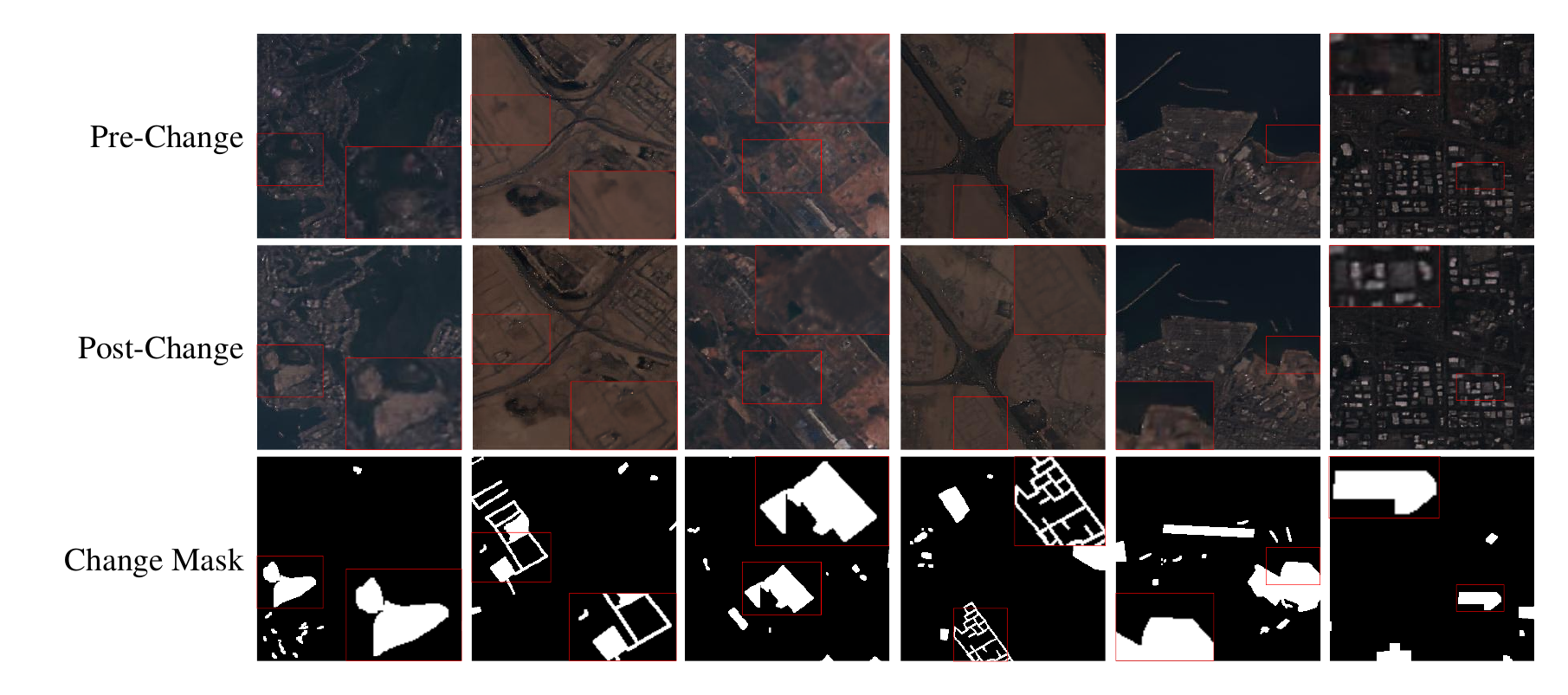}
    \vspace{-4mm}
    \caption{\textbf{Generated samples-OSCD.} We visualize several samples generated by our method and highlight the changed regions. The masks can annotate these changes in high accuracy.}
    \label{fig:CD samples}
\end{figure*}

\noindent
\textbf{Change Detection Dataset Synthesis.}
We further evaluate the visual quality of our synthetic data for change detection. The FID scores are displayed in Table \ref{tab:visual quality-cd}. Compared to LDM, our generated samples have lower FID scores and are closer to the real distribution. Qualitative samples generated by our method are shown in Fig. \ref{fig:CD samples}. The obvious changed area are highlighted. Comparing these regions, we can observe that the generator indeed generated changed images. Moreover, the generated masks annotated these changes in high accuracy.

\begin{table}[t]
  \centering
  \resizebox{\linewidth}{!}{
    \begin{tabular}{lccc}
    \toprule
    Method & FID (Image$_1$)$\downarrow$ & FID (Image$_2$)$\downarrow$ & FID (Mask)$\downarrow$ \\
    \midrule
    LDM-CD & 40.21 & 45.99 & 11.90\\
    Ours-CD & 10.15 & 10.74 & 0.05\\
    \bottomrule
  \end{tabular}}
  \vspace{-3mm}
  \caption{\textbf{Quantitative evaluation of synthetic dataset-OSCD.} Our method outperforms LDM in both image and mask generation.}
  \label{tab:visual quality-cd}
\end{table}

\subsection{Downstream Task Evaluation}
\label{sec:downstream}

\begin{table*}[h]
\centering
    \subtable[Segmantic Segmentation]{
    \label{tab:downstream-ss}
    \resizebox{0.47\textwidth}{!}{
    \begin{tabular}{l|cc|cc}
    \hline
         Method &Real Data   &Synthetic Data  &mIoU$\uparrow$ &F1$\uparrow$\\
    \hline
        \multirow{3}*{PFSegNet-r50 \cite{pfsegnet}}    &2k &-  &0.3250 &0.4444\\
         ~  &-  &10k    &0.3193  &0.4327\\
         ~  &2k &10k    &\textbf{0.3763} &\textbf{0.5021}\\
     \hline
        \multirow{3}*{PFSegNet-r101 \cite{pfsegnet}}    &2k &-  &0.3654 &0.4943\\
         ~  &-  &10k    &0.2532  &0.3609\\
         ~  &2k &10k    &\textbf{0.3697} &\textbf{0.4987}\\
    \hline
        \multirow{3}*{SegFormer-B0 \cite{segformer}}    &2k &-  &0.3377 &0.4605\\
         ~  &-  &10k    &0.2724  &0.3879\\
         ~  &2k &10k    &\textbf{0.3512} &\textbf{0.4788}\\
    \hline
        \multirow{3}*{SegFormer-B5 \cite{segformer}}    &2k &-  &0.3574 &0.4852\\
         ~  &-  &10k    &0.2932  &0.4088\\
         ~  &2k &10k    &\textbf{0.3772} &\textbf{0.5092}\\
    \hline  
    \end{tabular}}
    }
    \hfill
    \subtable[Change Detection]{
    \label{tab:downstream-cd}
    \resizebox{0.48\textwidth}{!}{
    \begin{tabular}{l|cc|cc}
    \hline
         Method &Real Data   &Synthetic Data  &mIoU$\uparrow$ &F1$\uparrow$\\
    \hline
        \multirow{3}*{SiamCRNN-r50 \cite{siamcrnn}}    &100 &-  &0.5292 &0.5947\\
         ~  &-  &500    &0.5210  &0.5809\\
         ~  &100 &500    &\textbf{0.5680} &\textbf{0.6486}\\
    \hline
        \multirow{3}*{SiamCRNN-r101 \cite{siamcrnn}}    &100 &-  &0.5191 &0.5766\\
         ~  &-  &500    &0.5269  &0.5970\\
         ~  &100 &500    &\textbf{0.5702} &\textbf{0.6494}\\
    \hline
        \multirow{3}*{ChangeFormerV1 \cite{changeformer}}    &100 &-  &0.5090 &0.5541\\
         ~  &-  &500    &0.5310  &0.5903\\
         ~  &100 &500    &\textbf{0.5395} &\textbf{0.6022}\\
    \hline
        \multirow{3}*{ChangeFormerV3 \cite{changeformer}}    &100 &-  &0.5460 &0.6274\\
         ~  &-  &500    &0.5561  &0.6390\\
         ~  &100 &500    &\textbf{0.5733} &\textbf{0.6617}\\
    \hline
    \end{tabular}}
    }
\caption{\textbf{Downstream task evaluation results} of (a) semantic segmantion and (b) change detection. With the augmentation of our synthetic data, the performance of downstream tasks on all methods get improvement, highlighted in \textbf{Bold}.}
\label{tab:downstream}
\end{table*}

We perform the corresponding downstream task experiments: semantic segmentation and change detection respectively, to further validate the effectiveness of our generated dataset.

\noindent
\textbf{Semantic Segmentation.} We limit the size of real dataset (using 2k pairs) and utilize 10k synthetic pairs to augment it. Then we train SS algorithms on these different dataset configuration (real data only, synthetic data only and augmented data) and evaluate on the same test set (real data). The mIoU and F1 score are used to evaluate the performance of the task. Table \ref{tab:downstream}(a) shows the segmentation results of all methods in different data configuration. As can be seen, without the supervised training of real data, the performance of SS algorithms will degrade. However, after augmenting the original real data with synthetic data, both SS models can achieve better results. Since the dataset obtained by the generative model is still learned from the real dataset, using only synthetic data to train the downstream SS model does not guarantee that the model can learn more knowledge of the feature of images, compared to training only on real data. During testing, performance degradation occurs due to the distribution difference between synthetic set and test set. Under the premise of ensuring real data supervision, using synthetic data for augmentation can enable the model to learn more knowledge to achieve better performance.


\noindent
\textbf{Change Detection.} Table \ref{tab:downstream}(b) presents the change detection results of all methods on three training configuration. We limit the size of real data to 100 and utilize 500 synthetic samples for augmentation. Both CD models achieve the significant improvement with the data augmentation. Moreover, due to the small size of the training set, training with only synthetic datasets dose not cause much performance degradation for CD models. For ResNet-101 backbone, SiamCRNN \cite{siamcrnn} even achieves better results when training only on the synthetic data compared to training only on the real data.


\begin{table*}[t]
  \centering
    \begin{tabular}{l|ccc|ccc}
    \hline
    \multirow{2}*{Method} & \multicolumn{3}{c|}{Reconstruction Quality} & \multicolumn{3}{c}{Synthesis Quality}\\
    \cline{2-7}
    ~ & RMSE$\downarrow$ & SAD$\downarrow$ & Cross Entropy$\downarrow$ & FID (Image$_1$)$\downarrow$ & FID (Image$_2$)$\downarrow$ & FID (Mask)$\downarrow$ \\
    \hline
    Ours-SS w/o SAD loss & 0.072 & 0.217 & \textbf{0.031} & 21.05 & - & 15.12\\
    Ours-SS & \textbf{0.025} & \textbf{0.103} & 0.051 & \textbf{5.19} & - & \textbf{10.79}\\
    Ours-CD w/o SAD loss & 0.051 & 0.179 & 0.020 & 31.30 & 32.49 & 0.26 \\
    Ours-CD & \textbf{0.034} & \textbf{0.086} & \textbf{0.014} & \textbf{10.15} & \textbf{10.74} & \textbf{0.05}\\
    \hline
  \end{tabular}
  \caption{\textbf{Ablation study-SAD loss.} After eliminating the SAD loss term, the reconstruction quality degrades in the first-stage training, leading to the degradation of synthesis quality.}
  \label{tab:Ablation study-sad}
\end{table*}

\noindent
\textbf{Comparison with LDM.} We further compare the effectiveness of synthetic data generated by LDM and our method. We choose SegFormer-B5 \cite{segformer} and ChangeFormerV3 \cite{changeformer} training only on the real dataset as the baseline for SS and CD task, respectively. The dataset configuration is set to use 2k real data for SS and 100 real data for CD, and augmented with 5 times synthetic data. Table \ref{tab:downstream-ldm} presents the comparison results of two method on these two tasks. Our method outperforms LDM on both tasks, which demonstrates that our generated samples not only have better visual effects, but also more helpful in promoting downstream tasks.

\begin{table}[t]
    \centering
    \begin{tabular}{l|cc|cc}
    \hline
        \multirow{2}*{Method}&\multicolumn{2}{c|}{SS}&\multicolumn{2}{c}{CD}\\
        \cline{2-5}
        ~&mIoU$\uparrow$&F1$\uparrow$&mIoU$\uparrow$&F1$\uparrow$\\
    \hline
        Baseline & 0.3574 &0.4852 &0.5460 &0.6274\\
        LDM &0.3499 &0.4803 &0.5341 &0.5975\\
        Ours &\textbf{0.3772} &\textbf{0.5092} &\textbf{0.5733} &\textbf{0.6617}\\
    \hline
    \end{tabular}
    \caption{\textbf{Downstream tasks results comparing to LDM.} SegFormer-B5 \cite{segformer} and ChangeFormerV3 \cite{changeformer} are used as the baseline for SS and CD task. Our synthetic data has more promotion for the baseline.}
    \vspace{-0.2in}
    \label{tab:downstream-ldm}
\end{table}



\subsection{Ablation Study}

In this work, we propose to take the two-stream VAE to learn the latent representations of input HSIs and semantic annotations respectively. In experiments of Sec. \ref{sec:syn performance} and Sec. \ref{sec:downstream}, we have demonstrated the effectiveness of this approach, by comparing it with typical LDM method. We now assess the impact of SAD loss proposed in Eq. (\ref{eq:image loss}) on the reconstruction and synthesis quality, and the impact of the size of synthetic dataset on downstream tasks.

\noindent
\textbf{SAD Loss.} The SAD loss is utilized to ensure the spectral fidelity while reconstructing the HSIs in the first-stage training. We eliminate this term and use only $L_1$ loss as the reconstruction loss. Table \ref{tab:Ablation study-sad} displays the results of these two configurations. In the first-stage training, the reconstruction quality degrades much after eliminating the SAD term, especially for the SAD metric, which leads to the degradation of image synthesis quality. For the reconstruction of masks, SAD loss will not influence the parameter update of mask branch, hence the reconstruction and synthesis quality of mask is not largely affected.

\noindent
\textbf{Size of Synthetic Dataset.} In Sec. \ref{sec:downstream}, we set the size of the synthetic dataset to be 5 times that of the real dataset. We further explore the impact of more augmentation configurations. Same as Sec. \ref{sec:downstream}, we choose SegFormer-B5 \cite{segformer} and ChangeFormerV3 \cite{changeformer} as the baseline for SS and CD task, training on the real data only. We use $1\times$ and $3\times$ synthetic data for augmentation. The results are displayed in Table \ref{tab:Ablation Study-train size}. As can be seen, the performance improves as the size of synthetic set increases.

\begin{table}[t]
    \centering
    \begin{tabular}{l|cc|cc}
    \hline
        \multirow{2}*{Syn Data}&\multicolumn{2}{c|}{SS}&\multicolumn{2}{c}{CD}\\
        \cline{2-5}
        ~&mIoU$\uparrow$&F1$\uparrow$&mIoU$\uparrow$&F1$\uparrow$\\
    \hline
        Baseline &0.3574 &0.4852 &0.5460 &0.6274\\
        $\times1$ &0.3664 &0.4942 &0.5666  &0.6546\\
        $\times3$ &0.3757 &0.5086 &0.5694  &0.6561\\
        $\times5$ &\textbf{0.3772} &\textbf{0.5092} &\textbf{0.5733}  &\textbf{0.6617}\\
    \hline
    \end{tabular}
    \caption{\textbf{Ablation study-size of synthetic dataset.} SegFormer-B5 \cite{segformer} and ChangeFormerV3 \cite{changeformer} are used as the baseline for SS and CD task. We gradually add the size of synthetic dataset. Results have shown that the performance improves as the size of synthetic set increases.}
    \vspace{-0.1in}
    \label{tab:Ablation Study-train size}
\end{table}
\section{Conclusion}
\label{sec:conclusion}
Our work demonstrates the value and potential of using diffusion models to generate synthetic data in a context where hyperspectral images are scarce and annotation is expensive. By using a two-stream VAE to simultaneously compress images and labels into the latent space and learn their joint distribution, it is possible to generate high-dimensional spectral data with semantic annotations. We have designed our generative model for two of the most widely used dense prediction tasks in hyperspectral remote sensing images: semantic segmentation and change detection, which can generate high-quality HSIs and pixel-level semantic annotations automatically, and validated the effectiveness of our synthetic dataset on these tasks. In data-hunger circumstances, augmenting the traing set with synthetic data can bring positive impacts on models of downstream tasks. 

\noindent
\textbf{Limitation}. For generated annotations, we currently have no suitable method to verify their pixel-level alignments with genetated images without the reference of ground truth. The reliability of generated samples can only be verified by downstream tasks right now. We will continue to explore how to evaluate the reliability of generated samples in the future.


{
    \newpage
    \small
    \bibliographystyle{ieeenat_fullname}
    \bibliography{main}

\begin{thebibliography}{66}
\providecommand{\natexlab}[1]{#1}
\providecommand{\url}[1]{\texttt{#1}}
\expandafter\ifx\csname urlstyle\endcsname\relax
  \providecommand{\doi}[1]{doi: #1}\else
  \providecommand{\doi}{doi: \begingroup \urlstyle{rm}\Url}\fi

\bibitem[Adam et~al.(2010)Adam, Mutanga, and Rugege]{vegetation3}
Elhadi Adam, Onisimo Mutanga, and Denis Rugege.
\newblock Multispectral and hyperspectral remote sensing for identification and mapping of wetland vegetation: a review.
\newblock \emph{Wetlands Ecology and Management}, 18:\penalty0 281--296, 2010.

\bibitem[Bandara and Patel(2022)]{changeformer}
Wele Gedara~Chaminda Bandara and Vishal~M. Patel.
\newblock A transformer-based siamese network for change detection.
\newblock In \emph{IGARSS 2022 - 2022 IEEE International Geoscience and Remote Sensing Symposium}, pages 207--210, 2022.

\bibitem[Bedini(2017)]{earth2}
Enton Bedini.
\newblock The use of hyperspectral remote sensing for mineral exploration: A review.
\newblock \emph{Journal of Hyperspectral Remote Sensing}, 7\penalty0 (4):\penalty0 189--211, 2017.

\bibitem[Cai et~al.(2022{\natexlab{a}})Cai, Lin, Hu, Wang, Yuan, Zhang, Timofte, and Van~Gool]{cai2022mask}
Yuanhao Cai, Jing Lin, Xiaowan Hu, Haoqian Wang, Xin Yuan, Yulun Zhang, Radu Timofte, and Luc Van~Gool.
\newblock Mask-guided spectral-wise transformer for efficient hyperspectral image reconstruction.
\newblock In \emph{Proceedings of the IEEE/CVF Conference on Computer Vision and Pattern Recognition}, pages 17502--17511, 2022{\natexlab{a}}.

\bibitem[Cai et~al.(2022{\natexlab{b}})Cai, Lin, Lin, Wang, Zhang, Pfister, Timofte, and Van~Gool]{cai2022mst++}
Yuanhao Cai, Jing Lin, Zudi Lin, Haoqian Wang, Yulun Zhang, Hanspeter Pfister, Radu Timofte, and Luc Van~Gool.
\newblock Mst++: Multi-stage spectral-wise transformer for efficient spectral reconstruction.
\newblock In \emph{Proceedings of the IEEE/CVF Conference on Computer Vision and Pattern Recognition}, pages 745--755, 2022{\natexlab{b}}.

\bibitem[Chen et~al.(2020)Chen, Wu, Du, Zhang, and Wang]{siamcrnn}
Hongruixuan Chen, Chen Wu, Bo Du, Liangpei Zhang, and Le Wang.
\newblock Change detection in multisource vhr images via deep siamese convolutional multiple-layers recurrent neural network.
\newblock \emph{IEEE Transactions on Geoscience and Remote Sensing}, 58\penalty0 (4):\penalty0 2848--2864, 2020.

\bibitem[Chen et~al.(2007)Chen, Warner, and Campagna]{earth3}
Xianfeng Chen, Timothy~A Warner, and David~J Campagna.
\newblock Integrating visible, near-infrared and short-wave infrared hyperspectral and multispectral thermal imagery for geological mapping at cuprite, nevada.
\newblock \emph{Remote Sensing of Environment}, 110\penalty0 (3):\penalty0 344--356, 2007.

\bibitem[Chen et~al.(2016)Chen, Jiang, Li, Jia, and Ghamisi]{chen2016deep}
Yushi Chen, Hanlu Jiang, Chunyang Li, Xiuping Jia, and Pedram Ghamisi.
\newblock Deep feature extraction and classification of hyperspectral images based on convolutional neural networks.
\newblock \emph{IEEE Transactions on Geoscience and Remote Sensing}, 54\penalty0 (10):\penalty0 6232--6251, 2016.

\bibitem[Child(2020)]{Child2020VeryDV}
Rewon Child.
\newblock Very deep vaes generalize autoregressive models and can outperform them on images.
\newblock \emph{ArXiv}, abs/2011.10650, 2020.

\bibitem[Croitoru et~al.(2023)Croitoru, Hondru, Ionescu, and Shah]{dm1}
Florinel-Alin Croitoru, Vlad Hondru, Radu~Tudor Ionescu, and Mubarak Shah.
\newblock Diffusion models in vision: A survey.
\newblock \emph{IEEE Transactions on Pattern Analysis and Machine Intelligence}, 45\penalty0 (9):\penalty0 10850--10869, 2023.

\bibitem[Daudt et~al.(2018)Daudt, Le~Saux, Boulch, and Gousseau]{oscd}
Rodrigo~Caye Daudt, Bertr Le~Saux, Alexandre Boulch, and Yann Gousseau.
\newblock Urban change detection for multispectral earth observation using convolutional neural networks.
\newblock In \emph{IGARSS 2018 - 2018 IEEE International Geoscience and Remote Sensing Symposium}, pages 2115--2118, 2018.

\bibitem[Devaranjan et~al.(2020)Devaranjan, Kar, and Fidler]{devaranjan2020meta}
Jeevan Devaranjan, Amlan Kar, and Sanja Fidler.
\newblock Meta-sim2: Unsupervised learning of scene structure for synthetic data generation.
\newblock In \emph{Computer Vision--ECCV 2020: 16th European Conference, Glasgow, UK, August 23--28, 2020, Proceedings, Part XVII 16}, pages 715--733. Springer, 2020.

\bibitem[Esser et~al.(2021)Esser, Rombach, and Ommer]{esser2021taming}
Patrick Esser, Robin Rombach, and Bjorn Ommer.
\newblock Taming transformers for high-resolution image synthesis.
\newblock In \emph{Proceedings of the IEEE/CVF Conference on Computer Vision and Pattern Recognition}, pages 12873--12883, 2021.

\bibitem[Fu et~al.(2023)Fu, Sun, Zhang, Zhang, Ren, Jia, and Li]{scarse1}
Hang Fu, Genyun Sun, Li Zhang, Aizhu Zhang, Jinchang Ren, Xiuping Jia, and Feng Li.
\newblock Three-dimensional singular spectrum analysis for precise land cover classification from uav-borne hyperspectral benchmark datasets.
\newblock \emph{ISPRS Journal of Photogrammetry and Remote Sensing}, 203:\penalty0 115--134, 2023.

\bibitem[Gaidon et~al.(2016)Gaidon, Wang, Cabon, and Vig]{gaidon2016virtual}
Adrien Gaidon, Qiao Wang, Yohann Cabon, and Eleonora Vig.
\newblock Virtual worlds as proxy for multi-object tracking analysis.
\newblock In \emph{Proceedings of the IEEE Conference on Computer Vision and Pattern Recognition}, pages 4340--4349, 2016.

\bibitem[Gao et~al.(2021)Gao, Rasmussen, Kulits, Scheller, Greenberger, and Ehlmann]{Gao_2021_CVPR}
Angela~F. Gao, Brandon Rasmussen, Peter Kulits, Eva~L. Scheller, Rebecca Greenberger, and Bethany~L. Ehlmann.
\newblock Generalized unsupervised clustering of hyperspectral images of geological targets in the near infrared.
\newblock In \emph{Proceedings of the IEEE/CVF Conference on Computer Vision and Pattern Recognition (CVPR) Workshops}, pages 4294--4303, 2021.

\bibitem[Goodfellow et~al.(2014)Goodfellow, Pouget-Abadie, Mirza, Xu, Warde-Farley, Ozair, Courville, and Bengio]{gan1}
Ian Goodfellow, Jean Pouget-Abadie, Mehdi Mirza, Bing Xu, David Warde-Farley, Sherjil Ozair, Aaron Courville, and Yoshua Bengio.
\newblock Generative adversarial nets.
\newblock \emph{Advances in Neural Information Processing Systems}, 27, 2014.

\bibitem[Grau and Gastellu-Etchegorry(2013)]{physics}
Eloi Grau and Jean-Philippe Gastellu-Etchegorry.
\newblock Radiative transfer modeling in the earth--atmosphere system with dart model.
\newblock \emph{Remote Sensing of Environment}, 139:\penalty0 149--170, 2013.

\bibitem[Han et~al.(2023)Han, Zhang, Wang, Wang, Huang, Li, Wang, Chen, Li, Feng, et~al.]{challenge1}
Wei Han, Xiaohan Zhang, Yi Wang, Lizhe Wang, Xiaohui Huang, Jun Li, Sheng Wang, Weitao Chen, Xianju Li, Ruyi Feng, et~al.
\newblock A survey of machine learning and deep learning in remote sensing of geological environment: Challenges, advances, and opportunities.
\newblock \emph{ISPRS Journal of Photogrammetry and Remote Sensing}, 202:\penalty0 87--113, 2023.

\bibitem[He et~al.(2023)He, Yuan, Li, Xiao, Liu, Shen, and Zhang]{he2023spectral}
Jiang He, Qiangqiang Yuan, Jie Li, Yi Xiao, Denghong Liu, Huanfeng Shen, and Liangpei Zhang.
\newblock Spectral super-resolution meets deep learning: Achievements and challenges.
\newblock \emph{Information Fusion}, 97:\penalty0 101812, 2023.

\bibitem[He et~al.(2017)He, Li, Liu, and Li]{he2017recent}
Lin He, Jun Li, Chenying Liu, and Shutao Li.
\newblock Recent advances on spectral--spatial hyperspectral image classification: An overview and new guidelines.
\newblock \emph{IEEE Transactions on Geoscience and Remote Sensing}, 56\penalty0 (3):\penalty0 1579--1597, 2017.

\bibitem[Heusel et~al.(2017)Heusel, Ramsauer, Unterthiner, Nessler, and Hochreiter]{fid}
Martin Heusel, Hubert Ramsauer, Thomas Unterthiner, Bernhard Nessler, and Sepp Hochreiter.
\newblock Gans trained by a two time-scale update rule converge to a local nash equilibrium.
\newblock \emph{Advances in Neural Information Processing Systems}, 30, 2017.

\bibitem[Ho et~al.(2020)Ho, Jain, and Abbeel]{dm2}
Jonathan Ho, Ajay Jain, and Pieter Abbeel.
\newblock Denoising diffusion probabilistic models.
\newblock \emph{Advances in Neural Information Processing Systems}, 33:\penalty0 6840--6851, 2020.

\bibitem[Hong et~al.(2024)Hong, Zhang, Li, Li, Li, Yao, Yokoya, Li, Ghamisi, Jia, Plaza, Gamba, Benediktsson, and Chanussot]{spectralgpt}
Danfeng Hong, Bing Zhang, Xuyang Li, Yuxuan Li, Chenyu Li, Jing Yao, Naoto Yokoya, Hao Li, Pedram Ghamisi, Xiuping Jia, Antonio Plaza, Paolo Gamba, Jon~Atli Benediktsson, and Jocelyn Chanussot.
\newblock Spectralgpt: Spectral remote sensing foundation model.
\newblock \emph{IEEE Transactions on Pattern Analysis and Machine Intelligence}, 46\penalty0 (8):\penalty0 5227--5244, 2024.

\bibitem[Im and Jensen(2008)]{vegetation2}
Jungho Im and John~R Jensen.
\newblock Hyperspectral remote sensing of vegetation.
\newblock \emph{Geography Compass}, 2\penalty0 (6):\penalty0 1943--1961, 2008.

\bibitem[Kang et~al.(2023)Kang, Zhu, Zhang, Park, Shechtman, Paris, and Park]{gan2}
Minguk Kang, Jun-Yan Zhu, Richard Zhang, Jaesik Park, Eli Shechtman, Sylvain Paris, and Taesung Park.
\newblock Scaling up gans for text-to-image synthesis.
\newblock In \emph{Proceedings of the IEEE/CVF Conference on Computer Vision and Pattern Recognition}, pages 10124--10134, 2023.

\bibitem[Kar et~al.(2019)Kar, Prakash, Liu, Cameracci, Yuan, Rusiniak, Acuna, Torralba, and Fidler]{kar2019meta}
Amlan Kar, Aayush Prakash, Ming-Yu Liu, Eric Cameracci, Justin Yuan, Matt Rusiniak, David Acuna, Antonio Torralba, and Sanja Fidler.
\newblock Meta-sim: Learning to generate synthetic datasets.
\newblock In \emph{Proceedings of the IEEE/CVF International Conference on Computer Vision}, pages 4551--4560, 2019.

\bibitem[Kim et~al.(2023)Kim, Liu, Jain, and Liu]{kim2023dcface}
Minchul Kim, Feng Liu, Anil Jain, and Xiaoming Liu.
\newblock Dcface: Synthetic face generation with dual condition diffusion model.
\newblock In \emph{Proceedings of the IEEE/CVF Conference on Computer Vision and Pattern Recognition}, pages 12715--12725, 2023.

\bibitem[Kirillov et~al.(2023)Kirillov, Mintun, Ravi, Mao, Rolland, Gustafson, Xiao, Whitehead, Berg, Lo, et~al.]{sam}
Alexander Kirillov, Eric Mintun, Nikhila Ravi, Hanzi Mao, Chloe Rolland, Laura Gustafson, Tete Xiao, Spencer Whitehead, Alexander~C Berg, Wan-Yen Lo, et~al.
\newblock Segment anything.
\newblock In \emph{Proceedings of the IEEE/CVF International Conference on Computer Vision}, pages 4015--4026, 2023.

\bibitem[Li et~al.(2022)Li, Ling, Kim, Kreis, Fidler, and Torralba]{bigdatasetgan}
Daiqing Li, Huan Ling, Seung~Wook Kim, Karsten Kreis, Sanja Fidler, and Antonio Torralba.
\newblock Bigdatasetgan: Synthesizing imagenet with pixel-wise annotations.
\newblock In \emph{Proceedings of the IEEE/CVF Conference on Computer Vision and Pattern Recognition}, pages 21330--21340, 2022.

\bibitem[Li et~al.(2019)Li, Song, Fang, Chen, Ghamisi, and Benediktsson]{li2019deep}
Shutao Li, Weiwei Song, Leyuan Fang, Yushi Chen, Pedram Ghamisi, and Jon~Atli Benediktsson.
\newblock Deep learning for hyperspectral image classification: An overview.
\newblock \emph{IEEE Transactions on Geoscience and Remote Sensing}, 57\penalty0 (9):\penalty0 6690--6709, 2019.

\bibitem[Li et~al.(2021)Li, He, Li, Li, Cheng, Shi, Weng, Tong, and Lin]{pfsegnet}
Xiangtai Li, Hao He, Xia Li, Duo Li, Guangliang Cheng, Jianping Shi, Lubin Weng, Yunhai Tong, and Zhouchen Lin.
\newblock Pointflow: Flowing semantics through points for aerial image segmentation.
\newblock In \emph{Proceedings of the IEEE/CVF Conference on Computer Vision and Pattern Recognition}, pages 4217--4226, 2021.

\bibitem[Li et~al.(2024)Li, Tucker, Snavely, and Holynski]{li2024generative}
Zhengqi Li, Richard Tucker, Noah Snavely, and Aleksander Holynski.
\newblock Generative image dynamics.
\newblock In \emph{Proceedings of the IEEE/CVF Conference on Computer Vision and Pattern Recognition}, pages 24142--24153, 2024.

\bibitem[Liu et~al.(2019)Liu, Marinelli, Bruzzone, and Bovolo]{liu2019review}
Sicong Liu, Daniele Marinelli, Lorenzo Bruzzone, and Francesca Bovolo.
\newblock A review of change detection in multitemporal hyperspectral images: Current techniques, applications, and challenges.
\newblock \emph{IEEE Geoscience and Remote Sensing Magazine}, 7\penalty0 (2):\penalty0 140--158, 2019.

\bibitem[Ma et~al.(2024)Ma, Wang, Jia, Chen, Liu, Li, Chen, and Qiao]{ma2024latte}
Xin Ma, Yaohui Wang, Gengyun Jia, Xinyuan Chen, Ziwei Liu, Yuan-Fang Li, Cunjian Chen, and Yu Qiao.
\newblock Latte: Latent diffusion transformer for video generation.
\newblock \emph{arXiv preprint arXiv:2401.03048}, 2024.

\bibitem[Manas et~al.(2021)Manas, Lacoste, Gir{\'o}-i Nieto, Vazquez, and Rodriguez]{manas2021seasonal}
Oscar Manas, Alexandre Lacoste, Xavier Gir{\'o}-i Nieto, David Vazquez, and Pau Rodriguez.
\newblock Seasonal contrast: Unsupervised pre-training from uncurated remote sensing data.
\newblock In \emph{Proceedings of the IEEE/CVF International Conference on Computer Vision}, pages 9414--9423, 2021.

\bibitem[Men et~al.(2020)Men, Mao, Jiang, Ma, and Lian]{Controllable_Person}
Yifang Men, Yiming Mao, Yuning Jiang, Wei-Ying Ma, and Zhouhui Lian.
\newblock Controllable person image synthesis with attribute-decomposed gan.
\newblock In \emph{2020 IEEE/CVF Conference on Computer Vision and Pattern Recognition (CVPR)}, pages 5083--5092, 2020.

\bibitem[Pan et~al.(2003)Pan, Healey, Prasad, and Tromberg]{face1}
Zhihong Pan, Glenn Healey, Manish Prasad, and Bruce Tromberg.
\newblock Face recognition in hyperspectral images.
\newblock \emph{IEEE Transactions on Pattern Analysis and Machine Intelligence}, 25\penalty0 (12):\penalty0 1552--1560, 2003.

\bibitem[Pang et~al.(2024)Pang, Tang, Xu, Meng, and Cao]{hsigene}
Li Pang, Datao Tang, Shuang Xu, Deyu Meng, and Xiangyong Cao.
\newblock Hsigene: A foundation model for hyperspectral image generation.
\newblock \emph{arXiv preprint arXiv:2409.12470}, 2024.

\bibitem[Peng et~al.(2021)Peng, Liu, Xu, and Li]{vae1}
Jialun Peng, Dong Liu, Songcen Xu, and Houqiang Li.
\newblock Generating diverse structure for image inpainting with hierarchical vq-vae.
\newblock In \emph{Proceedings of the IEEE/CVF Conference on Computer Vision and Pattern Recognition}, pages 10775--10784, 2021.

\bibitem[Qu et~al.(2024)Qu, Zhao, Dong, Xiao, Li, and Du]{10354413}
Jiahui Qu, Jingyu Zhao, Wenqian Dong, Song Xiao, Yunsong Li, and Qian Du.
\newblock Feature mutual representation-based graph domain adaptive network for unsupervised hyperspectral change detection.
\newblock \emph{IEEE Transactions on Geoscience and Remote Sensing}, 62:\penalty0 1--14, 2024.

\bibitem[Razavi et~al.(2019)Razavi, Van~den Oord, and Vinyals]{vae2}
Ali Razavi, Aaron Van~den Oord, and Oriol Vinyals.
\newblock Generating diverse high-fidelity images with vq-vae-2.
\newblock \emph{Advances in Neural Information Processing Systems}, 32, 2019.

\bibitem[Rombach et~al.(2022)Rombach, Blattmann, Lorenz, Esser, and Ommer]{ldm}
Robin Rombach, Andreas Blattmann, Dominik Lorenz, Patrick Esser, and Bj{\"o}rn Ommer.
\newblock High-resolution image synthesis with latent diffusion models.
\newblock In \emph{Proceedings of the IEEE/CVF Conference on Computer Vision and Pattern Recognition}, pages 10684--10695, 2022.

\bibitem[Ronneberger et~al.(2015)Ronneberger, Fischer, and Brox]{unet}
Olaf Ronneberger, Philipp Fischer, and Thomas Brox.
\newblock U-net: Convolutional networks for biomedical image segmentation.
\newblock In \emph{Medical Image Computing and Computer-Assisted Intervention--MICCAI 2015: 18th international conference, Munich, Germany, October 5-9, 2015, proceedings, part III 18}, pages 234--241. Springer, 2015.

\bibitem[Shi et~al.(2019)Shi, Paige, Torr, et~al.]{shi2019variational}
Yuge Shi, Brooks Paige, Philip Torr, et~al.
\newblock Variational mixture-of-experts autoencoders for multi-modal deep generative models.
\newblock \emph{Advances in Neural Information Processing Systems}, 32, 2019.

\bibitem[Shorten and Khoshgoftaar(2019)]{affine}
Connor Shorten and Taghi~M Khoshgoftaar.
\newblock A survey on image data augmentation for deep learning.
\newblock \emph{Journal of Big Data}, 6\penalty0 (1):\penalty0 1--48, 2019.

\bibitem[Sun and Du(2019)]{sun2019hyperspectral}
Weiwei Sun and Qian Du.
\newblock Hyperspectral band selection: A review.
\newblock \emph{IEEE Geoscience and Remote Sensing Magazine}, 7\penalty0 (2):\penalty0 118--139, 2019.

\bibitem[Thenkabail et~al.(2000)Thenkabail, Smith, and De~Pauw]{vegetation1}
Prasad~S Thenkabail, Ronald~B Smith, and Eddy De~Pauw.
\newblock Hyperspectral vegetation indices and their relationships with agricultural crop characteristics.
\newblock \emph{Remote Sensing of Environment}, 71\penalty0 (2):\penalty0 158--182, 2000.

\bibitem[Toker et~al.(2024)Toker, Eisenberger, Cremers, and Leal-Taix{\'e}]{satsynth}
Aysim Toker, Marvin Eisenberger, Daniel Cremers, and Laura Leal-Taix{\'e}.
\newblock Satsynth: Augmenting image-mask pairs through diffusion models for aerial semantic segmentation.
\newblock In \emph{Proceedings of the IEEE/CVF Conference on Computer Vision and Pattern Recognition}, pages 27695--27705, 2024.

\bibitem[Uzair et~al.(2015)Uzair, Mahmood, and Mian]{face2}
Muhammad Uzair, Arif Mahmood, and Ajmal Mian.
\newblock Hyperspectral face recognition with spatiospectral information fusion and pls regression.
\newblock \emph{IEEE Transactions on Image Processing}, 24\penalty0 (3):\penalty0 1127--1137, 2015.

\bibitem[Vahdat and Kautz(2020)]{vahdat2020nvae}
Arash Vahdat and Jan Kautz.
\newblock Nvae: A deep hierarchical variational autoencoder.
\newblock \emph{Advances in Neural Information Pprocessing Systems}, 33:\penalty0 19667--19679, 2020.

\bibitem[Van~der Meer et~al.(2012)Van~der Meer, Van~der Werff, Van~Ruitenbeek, Hecker, Bakker, Noomen, Van Der~Meijde, Carranza, De~Smeth, and Woldai]{earth1}
Freek~D Van~der Meer, Harald~MA Van~der Werff, Frank~JA Van~Ruitenbeek, Chris~A Hecker, Wim~H Bakker, Marleen~F Noomen, Mark Van Der~Meijde, E~John~M Carranza, J~Boudewijn De~Smeth, and Tsehaie Woldai.
\newblock Multi-and hyperspectral geologic remote sensing: A review.
\newblock \emph{International Journal of Applied Earth Observation and Geoinformation}, 14\penalty0 (1):\penalty0 112--128, 2012.

\bibitem[Verrelst et~al.(2015)Verrelst, Camps-Valls, Mu{\~n}oz-Mar{\'\i}, Rivera, Veroustraete, Clevers, and Moreno]{verrelst2015optical}
Jochem Verrelst, Gustau Camps-Valls, Jordi Mu{\~n}oz-Mar{\'\i}, Juan~Pablo Rivera, Frank Veroustraete, Jan~GPW Clevers, and Jos{\'e} Moreno.
\newblock Optical remote sensing and the retrieval of terrestrial vegetation bio-geophysical properties--a review.
\newblock \emph{ISPRS Journal of Photogrammetry and Remote Sensing}, 108:\penalty0 273--290, 2015.

\bibitem[Wang et~al.(2023)Wang, Zhang, Li, and Tao]{wang2023multistage}
Junjie Wang, Mengmeng Zhang, Wei Li, and Ran Tao.
\newblock A multistage information complementary fusion network based on flexible-mixup for hsi-x image classification.
\newblock \emph{IEEE Transactions on Neural Networks and Learning Systems}, 2023.

\bibitem[Wang et~al.(2019)Wang, Yang, Li, Liang, Zhang, Hall, and Hu]{Example-Guided}
Miao Wang, Guo-Ye Yang, Ruilong Li, Run-Ze Liang, Song-Hai Zhang, Peter~M. Hall, and Shi-Min Hu.
\newblock Example-guided style-consistent image synthesis from semantic labeling.
\newblock In \emph{2019 IEEE/CVF Conference on Computer Vision and Pattern Recognition (CVPR)}, pages 1495--1504, 2019.

\bibitem[Wu et~al.(2023{\natexlab{a}})Wu, Zhao, Chen, Gu, Zhao, He, Zhou, Shou, and Shen]{datasetdm}
Weijia Wu, Yuzhong Zhao, Hao Chen, Yuchao Gu, Rui Zhao, Yefei He, Hong Zhou, Mike~Zheng Shou, and Chunhua Shen.
\newblock Datasetdm: Synthesizing data with perception annotations using diffusion models.
\newblock \emph{Advances in Neural Information Processing Systems}, 36:\penalty0 54683--54695, 2023{\natexlab{a}}.

\bibitem[Wu et~al.(2023{\natexlab{b}})Wu, Zhao, Shou, Zhou, and Shen]{diffumask}
Weijia Wu, Yuzhong Zhao, Mike~Zheng Shou, Hong Zhou, and Chunhua Shen.
\newblock Diffumask: Synthesizing images with pixel-level annotations for semantic segmentation using diffusion models.
\newblock In \emph{Proceedings of the IEEE/CVF International Conference on Computer Vision}, pages 1206--1217, 2023{\natexlab{b}}.

\bibitem[Xie et~al.(2021)Xie, Wang, Yu, Anandkumar, Alvarez, and Luo]{segformer}
Enze Xie, Wenhai Wang, Zhiding Yu, Anima Anandkumar, Jose~M Alvarez, and Ping Luo.
\newblock Segformer: Simple and efficient design for semantic segmentation with transformers.
\newblock \emph{Advances in Neural Information Processing Systems}, 34:\penalty0 12077--12090, 2021.

\bibitem[Xu et~al.(2023)Xu, Bai, Yu, Chang, Atkinson, and Ghamisi]{challenge2}
Yonghao Xu, Tao Bai, Weikang Yu, Shizhen Chang, Peter~M Atkinson, and Pedram Ghamisi.
\newblock Ai security for geoscience and remote sensing: Challenges and future trends.
\newblock \emph{IEEE Geoscience and Remote Sensing Magazine}, 11\penalty0 (2):\penalty0 60--85, 2023.

\bibitem[Yu et~al.(2024{\natexlab{a}})Yu, Pan, Ma, Mei, Chen, and Ma]{yu2024unmixdiff}
Yang Yu, Erting Pan, Yong Ma, Xiaoguang Mei, Qihai Chen, and Jiayi Ma.
\newblock Unmixdiff: Unmixing-based diffusion model for hyperspectral image synthesis.
\newblock \emph{IEEE Transactions on Geoscience and Remote Sensing}, 2024{\natexlab{a}}.

\bibitem[Yu et~al.(2024{\natexlab{b}})Yu, Pan, Wang, Wu, Mei, and Ma]{yu2024unmixing}
Yang Yu, Erting Pan, Xinya Wang, Yuheng Wu, Xiaoguang Mei, and Jiayi Ma.
\newblock Unmixing before fusion: A generalized paradigm for multi-source-based hyperspectral image synthesis.
\newblock In \emph{Proceedings of the IEEE/CVF Conference on Computer Vision and Pattern Recognition}, pages 9297--9306, 2024{\natexlab{b}}.

\bibitem[Zhang et~al.(2023{\natexlab{a}})Zhang, Rao, and Agrawala]{zhang2023adding}
Lvmin Zhang, Anyi Rao, and Maneesh Agrawala.
\newblock Adding conditional control to text-to-image diffusion models.
\newblock In \emph{Proceedings of the IEEE/CVF International Conference on Computer Vision}, pages 3836--3847, 2023{\natexlab{a}}.

\bibitem[Zhang et~al.(2023{\natexlab{b}})Zhang, Wang, Wang, Gong, Wu, and Li]{zhang2023features}
Mingyang Zhang, Zhaoyang Wang, Xiangyu Wang, Maoguo Gong, Yue Wu, and Hao Li.
\newblock Features kept generative adversarial network data augmentation strategy for hyperspectral image classification.
\newblock \emph{Pattern Recognition}, 142:\penalty0 109701, 2023{\natexlab{b}}.

\bibitem[Zhang and Zhao(2021)]{face3}
Xianyi Zhang and Haitao Zhao.
\newblock Hyperspectral-cube-based mobile face recognition: A comprehensive review.
\newblock \emph{Information Fusion}, 74:\penalty0 132--150, 2021.

\bibitem[Zhang et~al.(2021)Zhang, Ling, Gao, Yin, Lafleche, Barriuso, Torralba, and Fidler]{datasetgan}
Yuxuan Zhang, Huan Ling, Jun Gao, Kangxue Yin, Jean-Francois Lafleche, Adela Barriuso, Antonio Torralba, and Sanja Fidler.
\newblock Datasetgan: Efficient labeled data factory with minimal human effort.
\newblock In \emph{Proceedings of the IEEE/CVF Conference on Computer Vision and Pattern Recognition}, pages 10145--10155, 2021.

\bibitem[Zhong et~al.(2020)Zhong, Hu, Luo, Wang, Zhao, and Zhang]{scarse2}
Yanfei Zhong, Xin Hu, Chang Luo, Xinyu Wang, Ji Zhao, and Liangpei Zhang.
\newblock Whu-hi: Uav-borne hyperspectral with high spatial resolution (h2) benchmark datasets and classifier for precise crop identification based on deep convolutional neural network with crf.
\newblock \emph{Remote Sensing of Environment}, 250:\penalty0 112012, 2020.

\end{thebibliography}
}


\end{document}